\documentclass[sigconf]{acmart}
\usepackage{subfigure}
\usepackage{booktabs}
\usepackage{multirow}
\usepackage{color}

\AtBeginDocument{%
  }

\copyrightyear{2023}
\acmYear{2023}
\setcopyright{acmlicensed}
\acmConference[MM '23]{Proceedings of the 31st ACM International Conference on Multimedia}{October 29-November 3, 2023}{Ottawa, ON, Canada}
\acmBooktitle{Proceedings of the 31st ACM International Conference on Multimedia (MM '23), October 29-November 3, 2023, Ottawa, ON, Canada}
\acmPrice{15.00}
\acmDOI{10.1145/3581783.3611873}
\acmISBN{979-8-4007-0108-5/23/10}

\acmSubmissionID{819}

\begin{document}

\title{Visual Causal Scene Refinement for Video Question Answering}

\author{Yushen Wei}
\authornote{Both authors contributed equally to this research.}
\email{weiysh8@mail2.sysu.edu.cn}
\orcid{0000-0002-0527-5463}
\affiliation{%
  \institution{Sun Yat-sen University}
  \country{China}
}

\author{Yang Liu}
\authornotemark[1]
\email{liuy856@mail.sysu.edu.cn}
\orcid{0000-0002-9423-9252}
\affiliation{%
  \institution{Sun Yat-sen University}
    \country{China}
}

\author{Hong Yan}
\email{yanh36@mail2.sysu.edu.cn}
\orcid{0000-0003-4100-6751}
\affiliation{%
  \institution{Sun Yat-sen University}
    \country{China}
}

\author{Guanbin Li}
\email{liguanbin@mail.sysu.edu.cn}
\orcid{0000-0002-4805-0926}
\affiliation{%
  \institution{Sun Yat-sen University}
    \country{China}
}

\author{Liang Lin}
\authornote{Corresponding author}
\email{linliang@ieee.org }
\orcid{0000-0003-2248-3755}
\affiliation{%
  \institution{Sun Yat-sen University}
    \country{China}
}

\renewcommand{\shortauthors}{Yushen Wei, Yang Liu, Hong Yan, Guanbin Li, \& Liang Lin}

\begin{abstract}
Existing methods for video question answering (VideoQA) often suffer from spurious correlations between different modalities, leading to a failure in identifying the dominant visual evidence and the intended question. Moreover, these methods function as black boxes, making it difficult to interpret the visual scene during the QA process. In this paper, to discover critical video segments and frames that serve as the visual causal scene for generating reliable answers, we present a causal analysis of VideoQA and propose a framework for cross-modal causal relational reasoning, named Visual Causal Scene Refinement (VCSR). Particularly, a set of causal front-door intervention operations is introduced to explicitly find the visual causal scenes at both segment and frame levels. Our VCSR involves two essential modules: i) the Question-Guided Refiner (QGR) module, which refines consecutive video frames guided by the question semantics to obtain more representative segment features for causal front-door intervention; ii) the Causal Scene Separator (CSS) module, which discovers a collection of visual causal and non-causal scenes based on the visual-linguistic causal relevance and estimates the causal effect of the scene-separating intervention in a contrastive learning manner. Extensive experiments on the NExT-QA, Causal-VidQA, and MSRVTT-QA datasets demonstrate the superiority of our VCSR in discovering visual causal scene and achieving robust video question answering. The code is available at \url{https://github.com/YangLiu9208/VCSR}.
\end{abstract}

\ccsdesc[500]{Computing methodologies~Causal reasoning and diagnostics}
\ccsdesc[500]{Computing methodologies~Temporal reasoning}

\keywords{Video Question Answering, Causal Reasoning, Cross-Modal.}
\begin{teaserfigure}
    \centering
    \subfigure[Explanation of VideoQA based on causal scene sets. \label{fig: causal scene}]{    \includegraphics[width=0.47\textwidth]{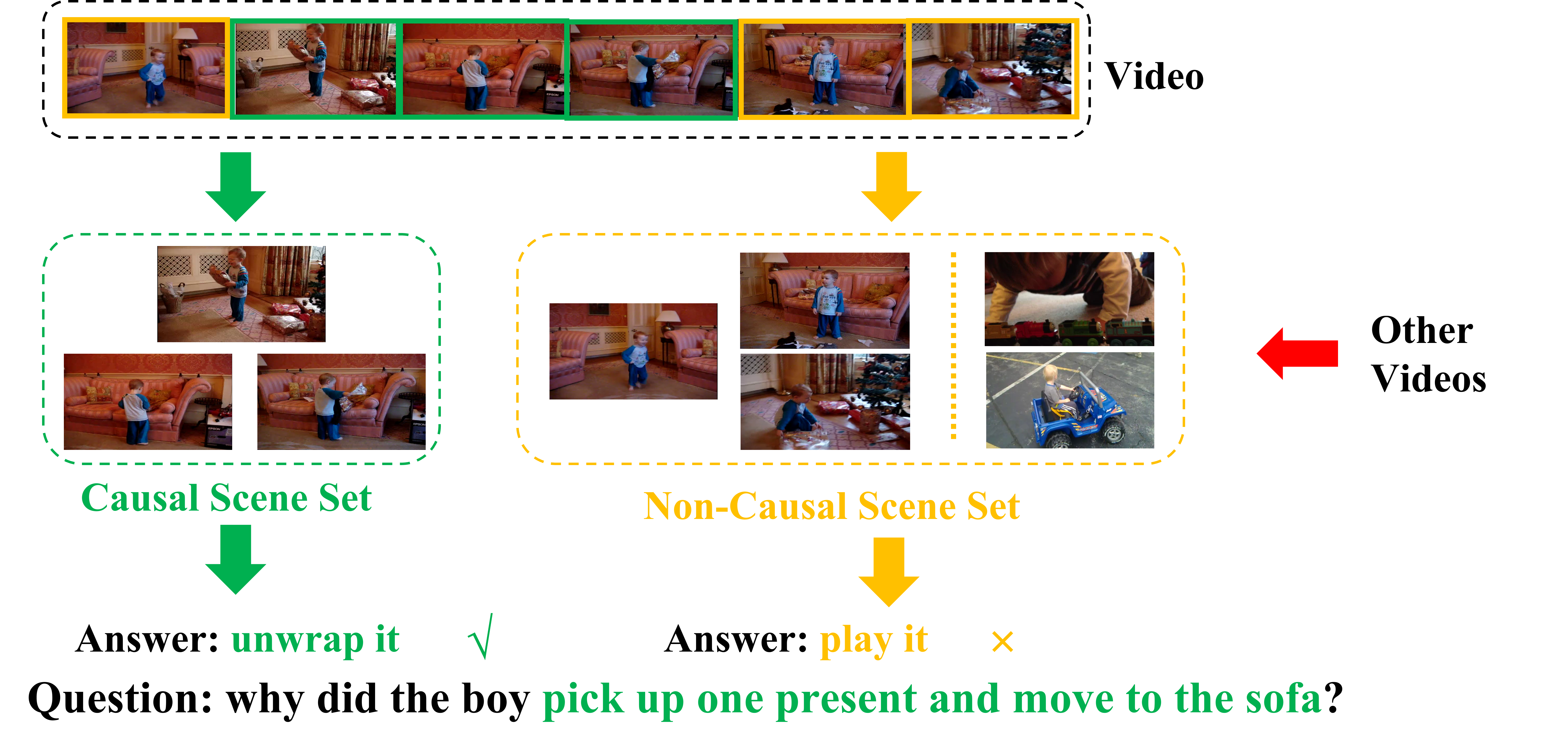}
    }
    \subfigure[Spurious correlations of visual contents in VideoQA tasks. \label{fig: spurious correlations}]{    \includegraphics[width=0.47\textwidth]{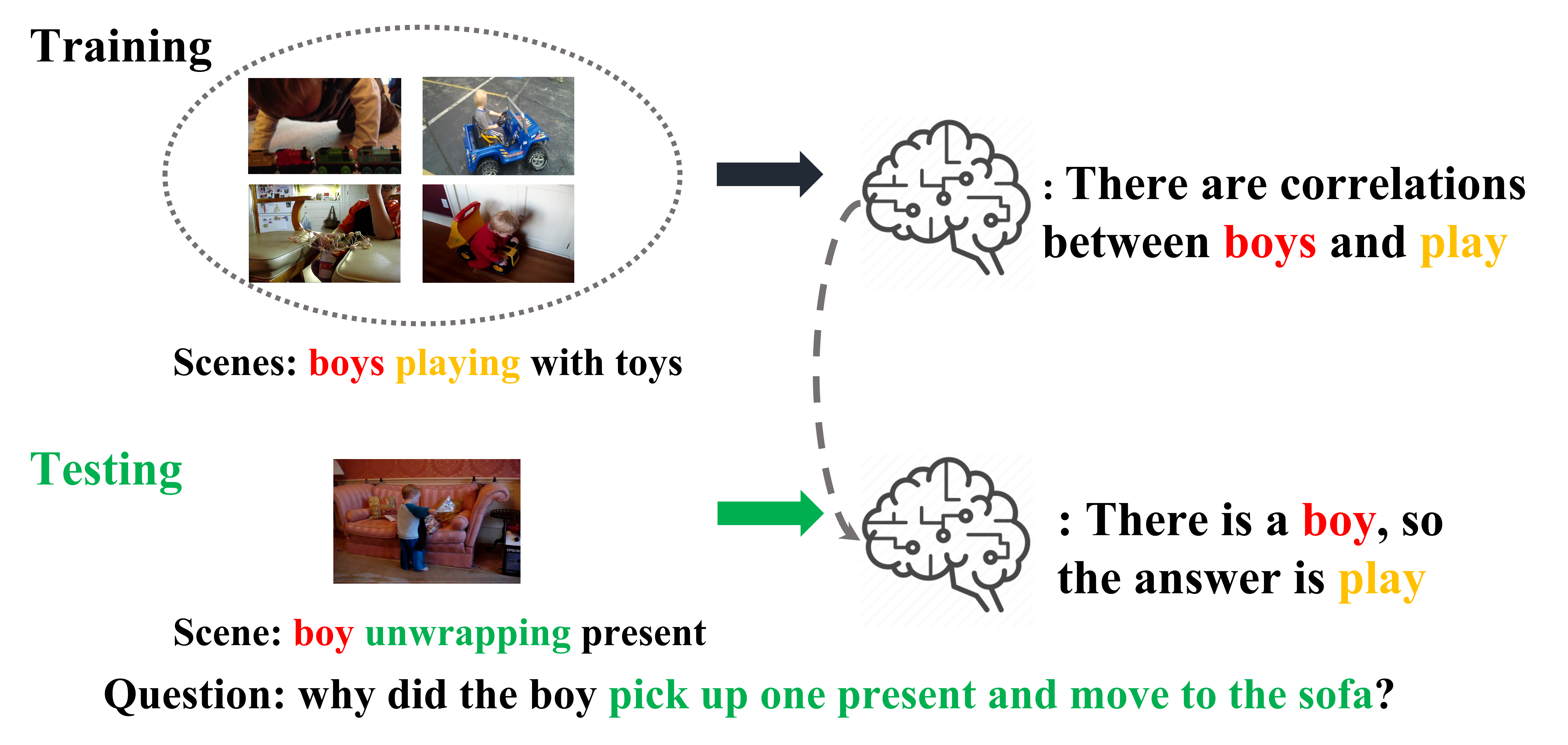}
    }
    \setlength{\abovecaptionskip}{0.2cm}
    \vspace{-10pt}
    \caption{An example of causal explanation of VideoQA. (a) illustrates the explanation of the model-predicted answer through a causal scene set, and (b) shows how the spurious correlation affects the model prediction.}
    \label{fig: Causal_explation}
\end{teaserfigure}

\maketitle

\section{Introduction}
Video question answering  \cite{le2020hierarchical,lei2021less} is a challenging task requiring machines to understand and interpret complex visual scenes to answer natural language questions about the content of a given video. Since videos have good potential to understand event temporality, causality, and dynamics, we focus on discovering question-critical visual causal scenes and achieving robust video question answering. Our task aims to fully comprehend the richer multi-modal event space and answer the given question in a causality-aware way. To achieve innovative architecture, several studies have explored VideoQA's multi-modal nature, including enhancing vision-language alignment \cite{jiang2020reasoning,park2021bridge} and reconsidering the structure of visual input \cite{le2020hierarchical,xiao2022video}. Most of the existing VideoQA methods \cite{li2019beyond,le2020hierarchical,park2021bridge} use recurrent neural networks (RNNs) \cite{sukhbaatar2015end}, attention mechanisms \cite{vaswani2017attention} or Graph Convolutional Networks \cite{kipf2016semi} for relation reasoning between visual and linguistic modalities. Although achieving promising results, the current video question answering methods suffer from two limitations.

First, the black-box nature of existing VideoQA models remains a significant challenge, as they lack transparency in their prediction process and offer little insight into the key visual cues used to answer questions about the video \cite{chen2020counterfactual,ross2017right,liu2023cross}. Specifically, it is difficult to explicitly discover the dominant visual segments or frames that the model focuses on to answer the question about the video. This lack of interpretability raises concerns about the robustness and reliability of the model, particularly in safety and security applications. To improve the interpretability of VideoQA models, it is crucial to identify a subset of visual scenes, referred to as ``causal scenes", that serve as evidence to support the answering process in a way that is interpretable to humans \cite{ross2017right}. For instance, Figure \ref{fig: causal scene}  shows that the causal scene set contains the boy's question-related action, which can serve as the dominant visual causal scene that provides an intuitive explanation for why the model gives the answer ``unwrap it". In contrast, the non-causal visual scene set includes question-irrelevant scenes that cannot faithfully reveal the correct question answering process.

Second, most of the existing video question answering models capture spurious visual correlations rather than the true causal structure, which leads to an unreliable reasoning process \cite{niu2021counterfactual,wang2021causal,li2022invariant,liu2022causal}. For instance, frequently co-occurring visual concepts, such as those illustrated in Figure \ref{fig: spurious correlations}, can be visual confounders ($C$). These confounders lead to a ``visual bias" denoting the strong correlations between visual features and answers. In the training set shown in Figure \ref{fig: spurious correlations}, the co-occurrence of the concepts ``boy" and ``play" dominates, which could lead the predictor to learn the spurious correlation between the two without considering the boy's action (i.e., causal positive scene $P$) to understand what the boy actually did. Consequently, there are significant differences in visual correlations between the training and testing sets, and memorizing strong visual priors can limit the reasoning ability of video question answering models. To mitigate visual spurious correlations, this paper takes a causal perspective on VideoQA by partitioning visual scenes into two parts: 1) causal positive scene $P$, which contains question-critical information, and 2) non-causal scene $N$, which is irrelevant to the answer. Thus, the non-causal scene $N$ is spuriously correlated with the answer $A$.

To address the aforementioned limitations, we propose the Visual Causal Scene Refinement (VCSR) framework to explicitly discover the visual causal scenes through causal front-door interventions. To obtain representative segment features for front-door intervention, we introduce the Question-Guided Refiner (QGR) module that refines consecutive video frames based on the question semantics. To identify visual causal and non-causal scenes, we propose the Causal Scene Separator (CSS) module based on the visual-linguistic causal relevance and estimates the causal effect of the scene-separating intervention through contrastive learning. Extensive experiments on the NExT-QA, Causal-VidQA, and MSRVTT-QA datasets demonstrate the superiority of VCSR over the state-of-the-art methods. Our main contributions are summarized as:
\begin{itemize}
\setlength{\itemsep}{0pt}
\setlength{\parsep}{0pt}
\setlength{\parskip}{0pt}
\item {\texttt{}We propose the Visual Causal Scene Refinement (VCSR), to explicitly discover true causal visual scenes from the perspective of causal front-door intervention. To the best of our knowledge, we are the first to discover visual causal scenes for video question answering.}
\item {\texttt{}We build the Causal Scene Separator (CSS) module that learns to discover a collection of visual causal and non-causal scenes based on the visual-linguistic causal relevance and estimates the causal effect of the scene-separating intervention contrastively.}
\item {\texttt{}We introduce the Question-Guided Refiner (QGR) module that refines consecutive video frames guided by the question semantics to obtain more representative segment features for causal front-door intervention.}
\end{itemize}

\section{Related Work}
\subsection{Video Question Answering}
Compared with image-based visual question answering \cite{antol2015vqa,yang2016stacked,anderson2018bottom}, video question answering is much more challenging due to the additional temporal dimension. To address the VideoQA problem, the model must capture spatial-temporal and visual-linguistic relationships to infer the answer. To explore relational reasoning in VideoQA, Xu et al. \cite{xu2017video} proposed an attention mechanism to exploit the appearance and motion knowledge with the question as guidance. Jang et al. \cite{jang2017tgif,jang2019video} proposed a dual-LSTM-based method with both spatial and temporal attention, which used a large-scale VideoQA dataset named TGIF-QA. Later, some hierarchical attention and co-attention-based methods \cite{li2019beyond,fan2019heterogeneous,jiayincai2020feature} were proposed to learn appearance-motion and question-related multi-modal interactions. Le et al. \cite{le2020hierarchical} proposed the hierarchical conditional relation network (HCRN) to construct sophisticated structures for representation and reasoning over videos. Jiang et al. \cite{jiang2020reasoning} introduced the heterogeneous graph alignment (HGA) nework that aligns the inter- and intra-modality information for cross-modal reasoning. Huang et al. \cite{huang2020location} proposed a location-aware graph convolutional network to reason over detected objects. Lei et al. \cite{lei2021less} employed sparse sampling to build a transformer-based model named CLIPBERT, which achieved end-to-end video-and-language understanding. Liu et al. \cite{liu2021hair} proposed the hierarchical visual-semantic relational reasoning (HAIR) framework to perform hierarchical relational reasoning. However, these previous works tend to capture cross-modal spurious correlations within the videos and neglect interpreting the visual scene during the QA process. In contrast, we propose the Visual Causal Scene Refinement (VCSR) architecture to explicitly refine the visual causal scenes temporally.

\subsection{Visual Causality Learning}
Compared to conventional debiasing techniques \cite{wang2020devil}, causal inference \cite{pearl2016causal,yang2021deconfounded,liu2022causal} has shown potential in mitigating spurious correlations \cite{bareinboim2012controlling} and disentangling model effects \cite{besserve2020counterfactuals} to achieve better generalization. Counterfactual and causal inference are gaining increasing attention in several computer vision tasks, including visual explanations \cite{goyal2019counterfactual,wang2020scout}, scene graph generation \cite{chen2019counterfactual,tang2020unbiased}, image recognition \cite{wang2020visual,wang2021causal}, video analysis \cite{fang2019modularized,kanehira2019multimodal,nan2021interventional}, and vision-language tasks \cite{abbasnejad2020counterfactual,niu2021counterfactual,yang2021causal,liu2022cross,chen2023visual}. Specifically, Tang et al. \cite{tang2020long}, Zhang et al. \cite{zhang2020causal}, Wang et al. \cite{wang2020visual}, and Qi et al. \cite{qi2020two} computed the direct causal effect and mitigated the bias based on observable confounders. Counterfactual based solutions are also effective. For example, Agarwal et al. \cite{agarwal2020towards} proposed a counterfactual sample synthesising method based on GAN \cite{goodfellow2014generative}. Chen et al. \cite{chen2020counterfactual} replaced critical objects and critical words with a mask token and reassigned an answer to synthesize counterfactual QA pairs. Apart from sample synthesising, Niu et al. \cite{niu2021counterfactual} developed a counterfactual VQA framework that reduces multi-modality bias by using a causality approach named Natural Indirect Effect and Total Direct Effect to eliminate the mediator effect. Li et al. \cite{li2022invariant} proposed an Invariant Grounding for VideoQA (IGV) to force models to shield the answering process from the negative influence of spurious correlations. Li et al. \cite{li2022equivariant} introduced a self-interpretable VideoQA framework named Equivariant and Invariant Grounding VideoQA (EIGV). Liu et al. \cite{liu2023cross} proposed a Cross-Modal Causal RelatIonal Reasoning (CMCIR) model for disentangling the visual and linguistic spurious correlations. Differently, our VCSR aims for visual causal scene discovery, which requires fine-grained understanding of spatial-temporal and visual-linguistic causal dependencies. Moreover, our VCSR explicitly finds the question-critical visual scenes temporally through front-door causal interventions.

\section{Methodology}
\subsection{VideoQA in Causal Perspective}
\begin{figure}[t]
    \centering
    \includegraphics[width=0.7\linewidth]{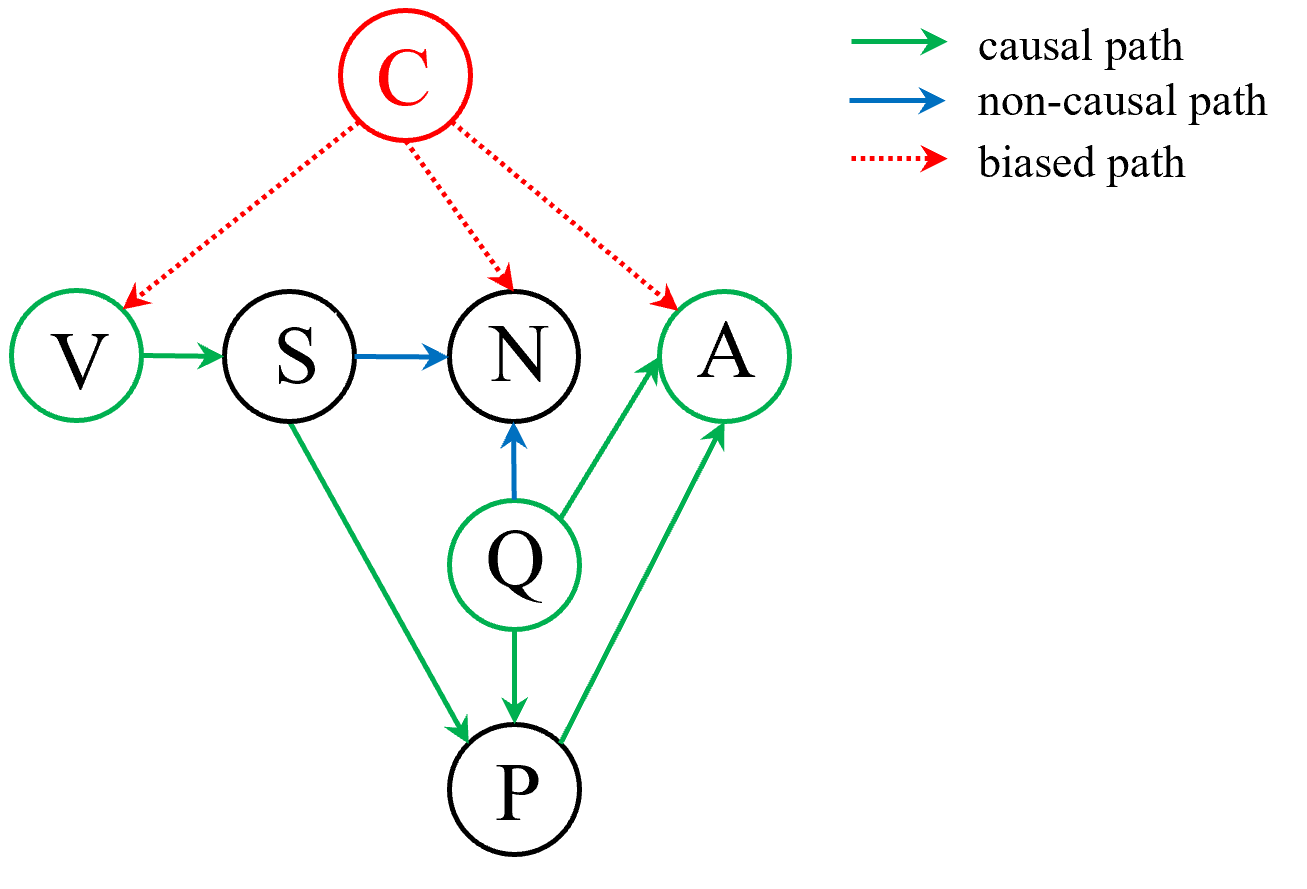}
    \vspace{-15pt}
    \caption{The Structured Causal Model (SCM) of VideoQA. V, Q and A denote video, question and answer respectively. C is the visual confounder, S denotes the refined video scenes, P and N are causal positive and negative visual scenes. \textcolor[rgb]{0, 0.69, 0.31}{Green flows}: the causal path of VideoQA (the front-door path). \textcolor[rgb]{0, 0.44, 0.75}{Blue flows}: the non-causal path. \textcolor{red}{Red flows}: biased VideoQA caused by the confounders (the back-door path).}
    \vspace{-15pt}
    \label{fig:scm}
\end{figure}

To discover visual causal scenes for VideoQA task, we employ Pearl's structural causal model (SCM) \cite{pearl2016causal} to model the causal effect between video-question pairs and the answer, as shown in Figure \ref{fig:scm}. The variables $V$, $Q$, $A$ are defined as the video, question, and answer. $S$ is refined video scene set which can be divided into causal positive scene set $P$  and negative scene set $N$. The front-door paths \emph{$V \rightarrow S \rightarrow P \rightarrow A$}, \emph{$Q \rightarrow P \rightarrow A$}, \emph{$Q\rightarrow A$} represent the true causal effects of VideoQA. These paths are involved in the reasoning process of watching the video, finding question-related scenes, and answering the question.  However, the visual confounder $C$ introduces a backdoor path $V \leftarrow C \rightarrow A$, which creates a spurious correlation between the video and answer. Unfortunately, visual domains have complex data biases, and it can be difficult to distinguish between different types of confounders. As a result, the visual confounder $C$ cannot be observed. Since the causal positive scenes $P$ completely mediates all causal effects from $V$ to $A$, to address this issue and achieve the true visual causal effect of $V \rightarrow S \rightarrow P \rightarrow A$, we propose a causal front-door intervention by treating $P$ as the mediator. The front-door intervention could be formulated as:
\begin{equation}
    \begin{aligned}
        &P(A|do(V), Q)  = \sum_p P(p|do(V = v))P(A|do(P = p), Q) \\
        & = \sum_p \sum_s P(p|s) P(s|do(V = v)) P(A|do(P = p), Q)\\
        & = \sum_p \sum_s P(p|s)P(s|v)\sum_{v'} \sum_{s'} P(A|p, s', Q) P(s'|v')P(v')
        \end{aligned}
    \label{eq: causal explanation1}
\end{equation}
where $do(\cdot)$ is the \emph{do}-operator indicating the intervention operation, and $P(s|do(V=v)) = P(s|v)$ because there is only front-door path between $V$ and $S$, $v'$ and $s'$ denotes intervened videos and segment sets after $do(P = p)$. Since the total scene set $S$ is determined  given a video, we could eliminate $s$ and $s'$ from the eq.\ref{eq: causal explanation1}:
\begin{equation}
    \begin{aligned}
        &P(A|do(V), Q)  = \sum_p P(p|v)\sum_{v'}P(A|p, v', Q)P(v')
    \end{aligned}
    \label{eq: causal explanation2}
\end{equation}
This is the front-door adjustment on causal path $V \rightarrow S \rightarrow P \rightarrow A$. After the intervention, we could eliminate the non-causal effect of back-door path $V \leftarrow C \rightarrow A$, making the model focus on the real causal effect. In section.\ref{seg:overall}, we propose the implementation of the front-door intervention eq.\ref{eq: causal explanation2}.

\begin{figure*}[t]
    \centering
    \includegraphics[width=0.75\linewidth]{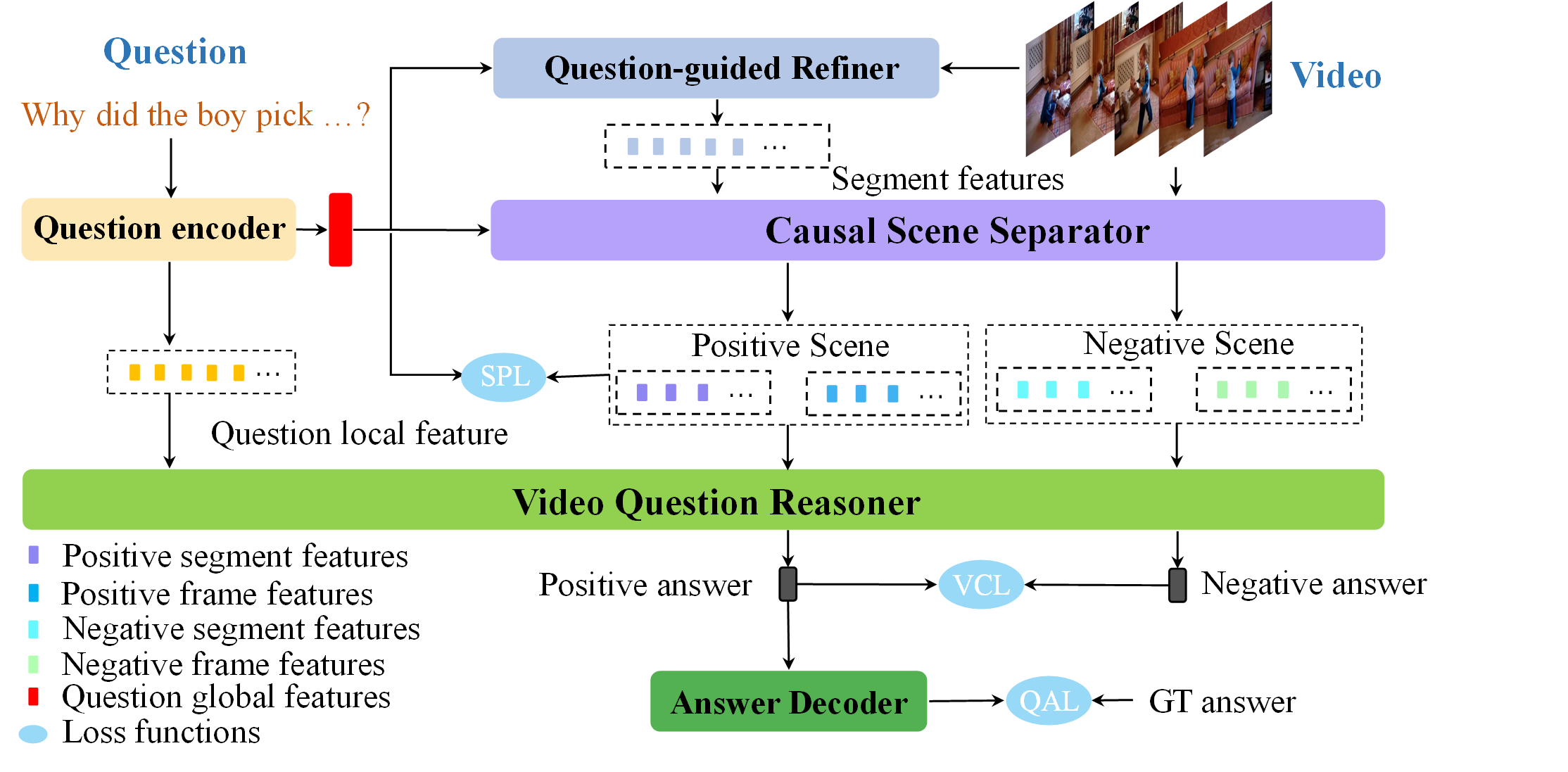}
    \vspace{-15pt}
    \caption{An overview of our Visual Causal Scene Refinement (VCSR) framework. The Question-Guided Refiner (QGR) encodes consecutive video frames guided by the question semantics to obtain representative segment features for causal front-door intervention. Then, the Causal Scene Separator (CSS) learns to construct a collection of visual causal and non-causal scenes based on the visual-linguistic causal relevance and estimates the causal effect of the scene-separating intervention in a contrastive learning manner. Finally, the Video Question Reasoner (VQR) computes the answer embedding with positive and negative video features. (SPL: Semantic Preserving Loss, VCL: Visual Contrastive Loss, QAL: Question Answering Loss) }
    \label{fig:overview}
        \vspace{-7pt}
\end{figure*}
\vspace{-5pt}

\subsection{Overall Causal Model Architecture}
\label{seg:overall}
To implement the front-door intervention, we propose: 1) a QGR (question-guided refiner) to construct the total scene set $S$ from video frames; 2) a CSS (causal scene separator) to model the causal positive scene distribution $P(p|v)$ in eq.\ref{eq: causal explanation2}, and a multi-modal transformer to parameterize the expectation of $P(A|p, v', Q)$; and 3) leverage a contrastive learning-based training objective to handle the causal intervention. The  framework of  VCSR is shown in Figure.\ref{fig:overview}.


\subsection{Question-Guided Refiner}
As shown in Figure.\ref{fig:scm} and eq.\ref{eq: causal explanation1}, the causal effect of video $V$ on the answer $A$ comes through the total scene set $S$. To construct the scene set $S$ from video segments, we designed Question-Guided Refiner (QGR) module to refine consecutive video frames by leveraging question semantics and obtaining more representative segment-level features for causal front-door intervention. Firstly, a pre-trained BERT model \cite{devlin2018bert} is employed to extract the question features from raw question texts. Next, the question features are encoded by a single-layer transformer encoder. The global representation of the question is denoted as the [CLS] features $q_g \in \mathbb{R} ^ \emph{d}$, while the concatenation of other output features represents the local question, denoted as $q_l$.

\begin{figure}[t]
    \centering
\includegraphics[width=1.0\linewidth]{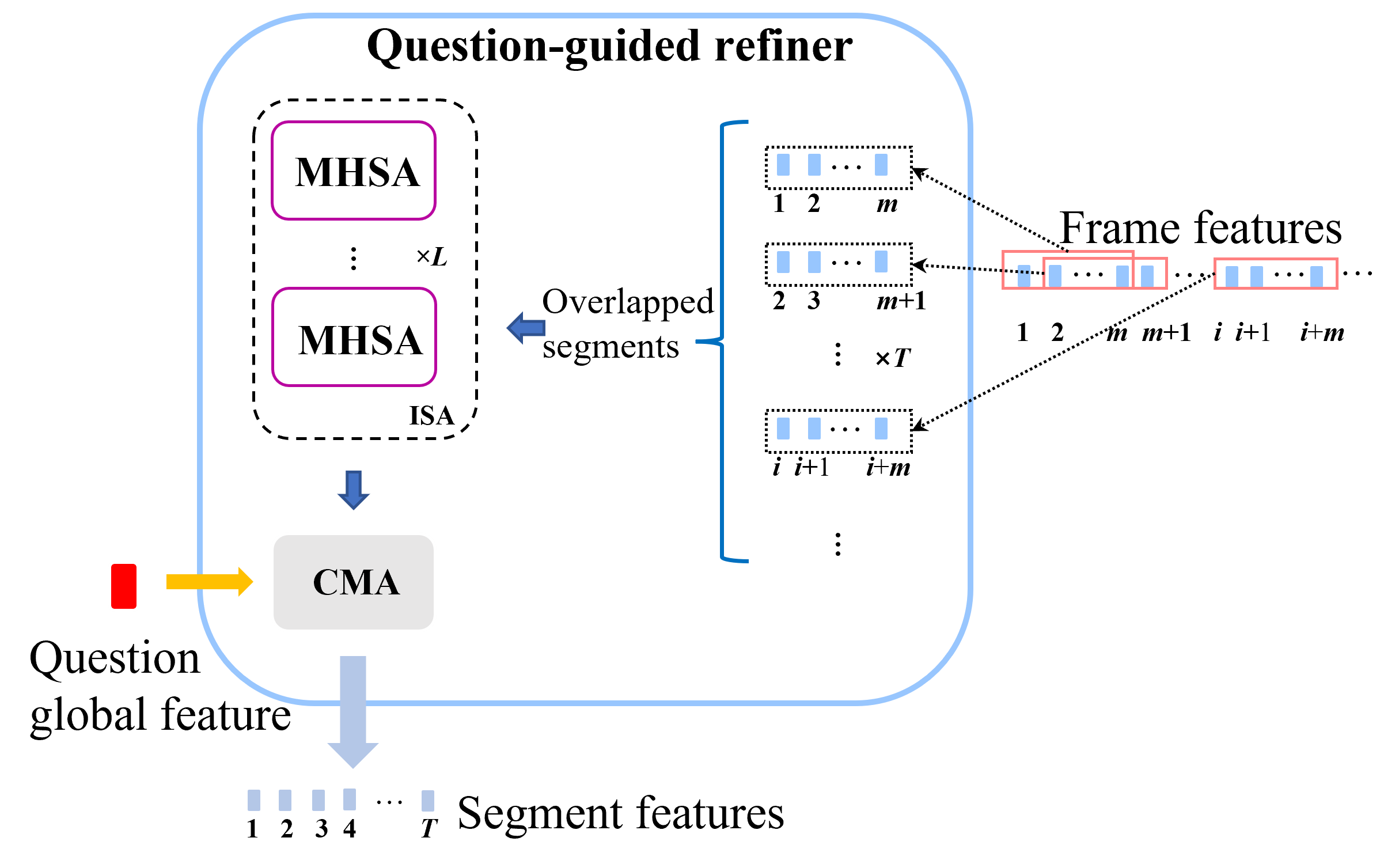}
\vspace{-25pt}
    \caption{The Question-Guided Refiner (QGR) module. The frame features are grouped into $T$ overlapped segments, then pass the In-segment Self Attention (ISA) module which contains $L$ layers of in-segment Multi-head Self Attention (MHSA), and finally question-guided Cross-modal Attention (CMA) aggregates frames in the same segment.}
    \vspace{-15pt}
    \label{fig:QGR}
\end{figure}

Given the original video $v$, we sparsely sample $N$ frames and utilize a pre-trained CLIP\cite{radford2021learning} encoder to extract the frame features $F_a = \{f_1, f_2, ... , f_N\}$, where $f_n \in \mathbb{R} ^ \emph{d}$, and $d$ denotes the dimension of the frame feature. Then, we combine $m$ adjacent frames to form a segment and obtain $T$ overlapping segments $S = \{s_1, s_2, ..., s_T\}$, where $s_t \in \mathbb{R} ^ {m \times d}$ denotes the frame features in a single segment, as Figure.\ref{fig:QGR} shows, each adjacent segments share $m-1$ overlapping frames. To mix the features within each segment, we employ an In-segment attention module (ISA), which is a transformer with $l$-layer multi-head self-attention module:
\vspace{-5pt}
\begin{equation}
    s'_t = [f'_{t, 1}, f'_{t, 2}, ... , f'_{t, m}]  =  \textrm{MHSA}^{(l)}(s_t + PE(s_t))
\end{equation}
where MHSA denotes the multi-head self-attention module, PE is the positional embedding and $f'_{i, j}$ is the $j$-th frame feature in the $i$-th segment.

The QGR module refines the frame features within the same segment to aggregate them temporally within the segment. To enhance the integration of feature aggregation with the VideoQA task, we incorporate a global question representation $q_g$ to guide our refining process. We begin by utilizing a cross-modal attention (CMA) module to obtain attention scores, which implicitly reflect the relevance of frames to the QA task. We then aggregate the frame features to refine the segment-level features using the attention scores obtained from the CMA module:
\begin{equation}
Q = f_q(q_g),~K = f_s(s'_t),~V = s'_t
\end{equation}
\begin{equation}
    s^*_t = \textrm{Softmax}(\frac{QK^T}{\sqrt{d_k}})V
\end{equation}
in which $f_q$ and $f_s$ are linear projection layers, and $s^*_t$ is the $t$-th segment feature after refining. Then, $T$ segment features are concatenated as the refined segments for the next causal scene separation step: $S^* = \{s^*_1, s^*_2, ..., s^*_T\}$, as shown in Figure.\ref{fig:QGR}.

\begin{figure*}
    \centering
    \subfigure[\label{fig:segment filter}]{
    \includegraphics[width=0.45\textwidth]{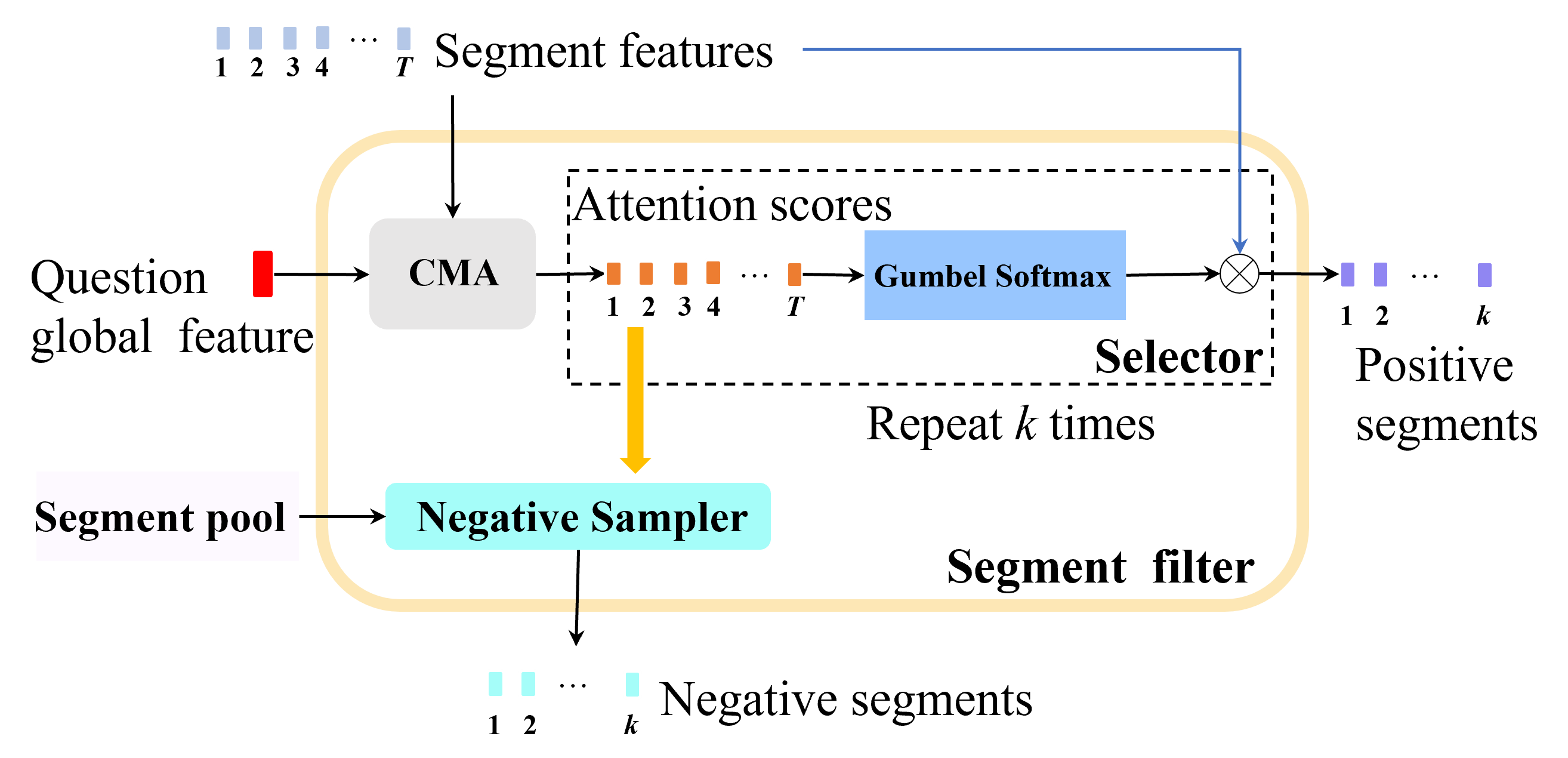}
    }
    \subfigure[\label{fig:frame filter}]{
    \includegraphics[width=0.45\textwidth]{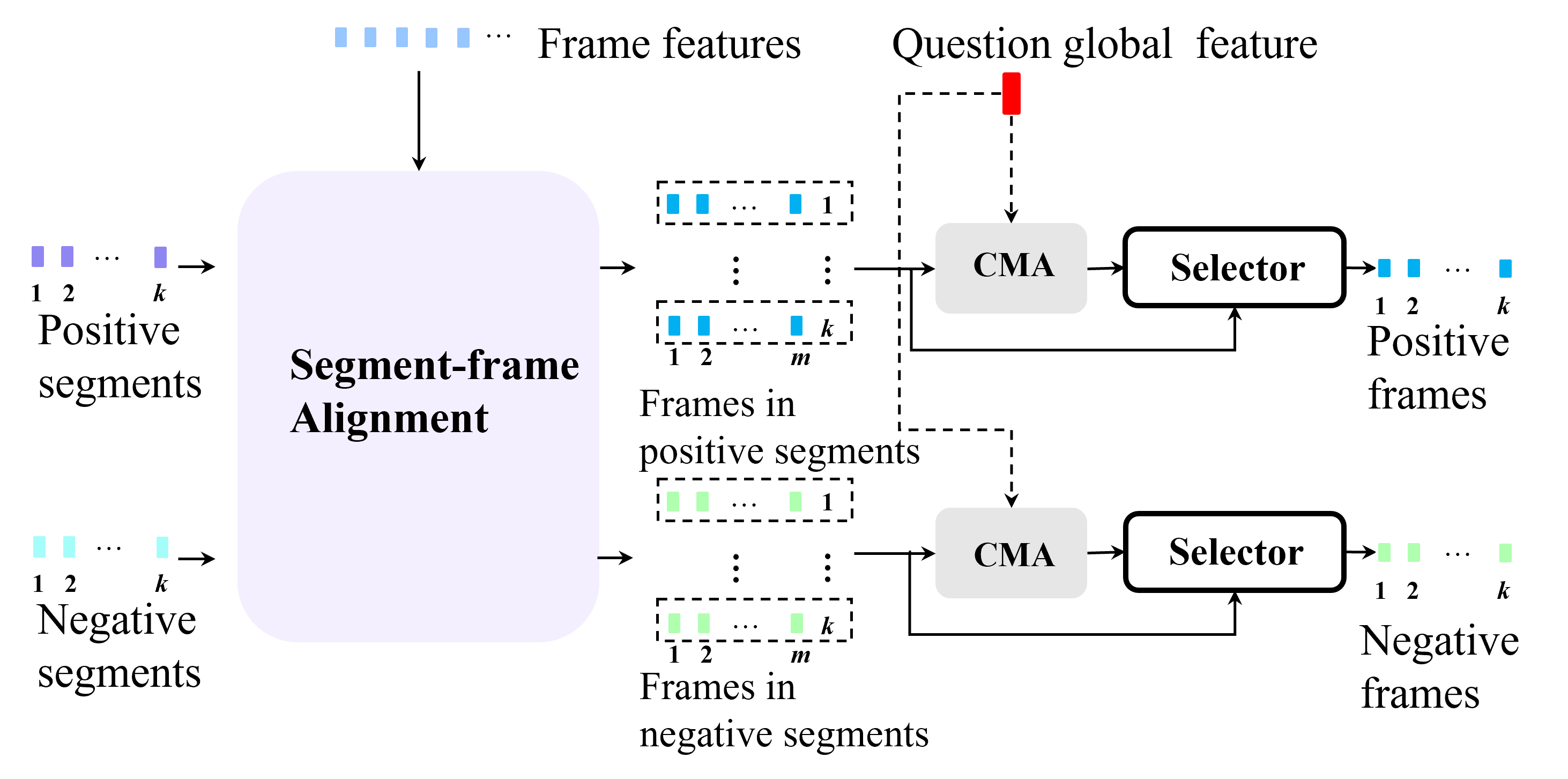}
    }
    \vspace{-15pt}
    \caption{The internal structure of Causal Scene Separator. (a) Causal segment generator. This module selects the possible causal positive segments in the video and generates negative segments by sampling segments from the segment pool. (b) Causal frame filter. Given positive and negative segments, the causal frame filter aligns segments with their respective frames, and then selects the most question-relevant frame for each positive and negative segment.}
    \vspace{-10pt}
    \label{fig: Causal Scene Seperator}
\end{figure*}

\subsection{Causal Scene Separator}
To construct a collection of causal scenes related to the question in a video(i.e., the positive scene $P$), we propose a Causal Scene Separator (CSS) that identifies segments and frames with high causal relevance to the question, as shown in Figure.\ref{fig: Causal Scene Seperator}. The Causal Scene Separator comprises two modules: a causal segment generator and a  causal frame filter.

\textbf{Causal segment generator}. The causal segment generator aims to generate sets of causal positive and negative segments for forming causal scenes. For positive segments, it initially computes the attention scores of the refined segments $S^*$ and the global question features $q_g$ using the cross-modal attention (CMA) module:
\begin{equation}
\vspace{-7pt}
    a_s = \textrm{Softmax}(g_q(q) \cdot g_s(S^*)^T)
\end{equation}
where $g_q$ and $g_s$ are linear layers. Then, we leverage Gumbel-Softmax to generate a discrete selection mask for capturing the causal content:
\begin{equation}
    s^i_p =  S^* \textrm{Gumbel-Softmax}(a_s)^T
\end{equation}
We repeat the selection process for $k$ times to obtain the causal positive segment set of size $k$, denoted as $S_p = \{s^1_p, s^2_p, ..., s^k_p\}$. Other segments with attentive probability lower than a threshold $\tau$ and segments from the segment pool, including segments from other videos, form a negative candidate set. A subset of the candidate set is sampled as the causal negative segment set $S_n = \{s^1_n, s^2_n,...,s^k_n\}$.

\begin{figure}[t]
    \centering
    \includegraphics[width = 0.9\linewidth]{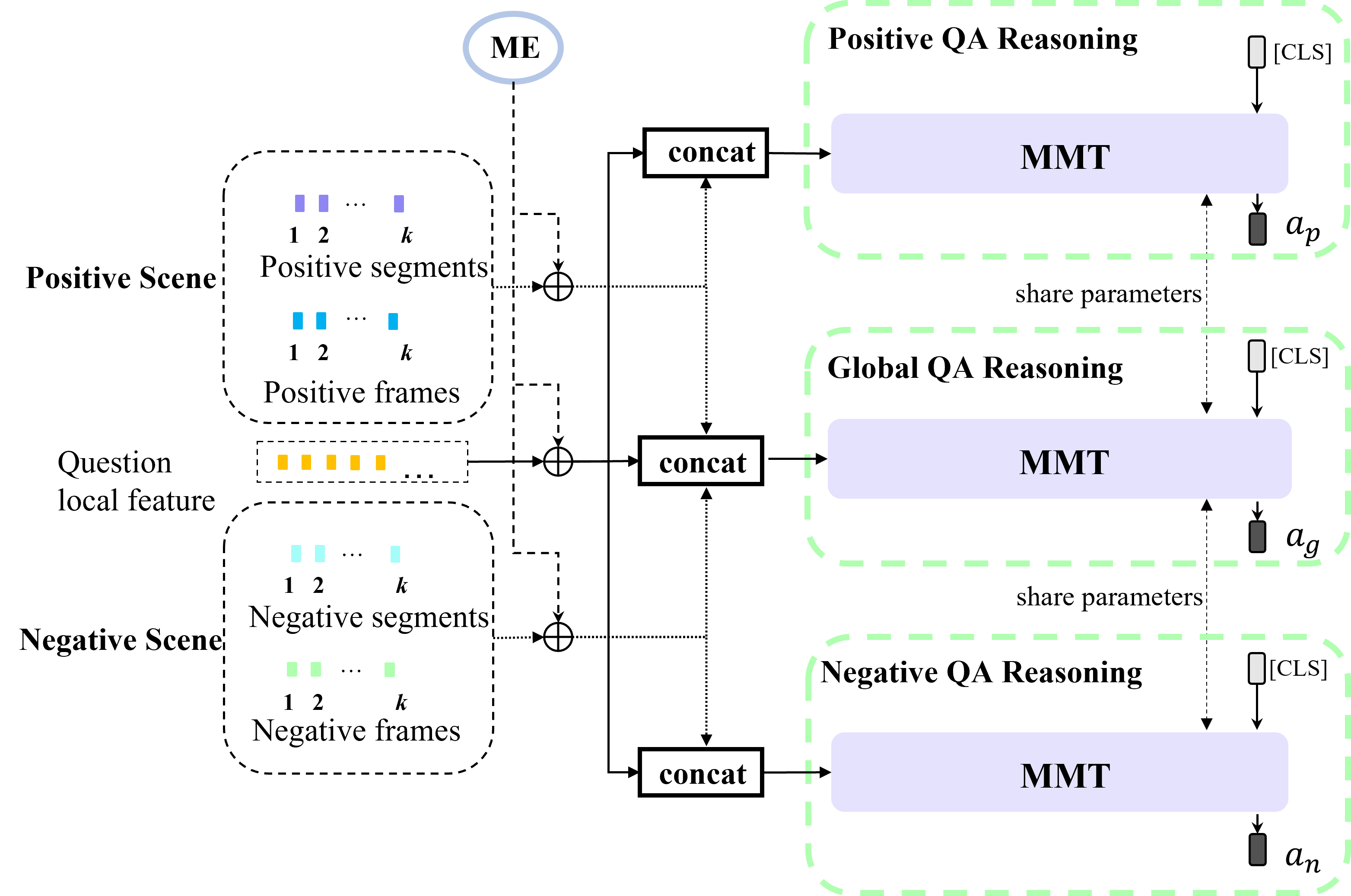}
    \vspace{-10pt}
    \caption{The multi-modal transformer (MMT) reasoner reasons the positive answer, negative answer, and global answer when given different causal scenes.}
    \vspace{-15pt}
    \label{fig:qa reasoner}
\end{figure}

\textbf{Causal frame filter}. Besides segment features, question-related frames in a segment can complement the causal scenarios since some causal scenes may only contain a few or even one single frame. As shown in Figure.\ref{fig:frame filter}, the causal frame filter first aligns the positive and negative segment sets with frames to obtain frames that belong to corresponding segments. Then, a selector similar to the one used in the segment filter in Figure.\ref{fig:segment filter} chooses a single frame for each segment to construct the causal positive frame set $F_p = \{f^1_p, f^2_p, ...,f^k_p\}$ and the causal negative frame set $F_n = \{f^1_n, f^2_n, ..., f^k_n\}$. The causal segment sets and the causal frame sets combine to form causal scenes. Formally, we have causal positive scene sets $C_p = \{S_p, F_p\}$ and causal negative scene sets $C_n = \{S_n, F_n\}$.

\begin{table*}
\resizebox{\textwidth}{!}{
\begin{tabular}{lccccc|cccc}
\hline
\multicolumn{1}{c}{\multirow{2}{*}{\textbf{Methods}}} & \multicolumn{1}{c}{\multirow{2}{*}{\textbf{Visual backbones}}} & \multicolumn{4}{c|}{\textbf{Val}}                                                                                                                          & \multicolumn{4}{c}{\textbf{Test}}                                                                                                                         \\ \cline{3-10}
\multicolumn{1}{c}{}                                  & \multicolumn{1}{c}{}                                           & \multicolumn{1}{l}{\textbf{Causal}} & \multicolumn{1}{l}{\textbf{Temporal}} & \multicolumn{1}{l}{\textbf{Descriptive}} & \multicolumn{1}{l|}{\textbf{Acc.}} & \multicolumn{1}{l}{\textbf{Causal}} & \multicolumn{1}{l}{\textbf{Temporal}} & \multicolumn{1}{l}{\textbf{Descriptive}} & \multicolumn{1}{l}{\textbf{Acc.}} \\ \hline
EVQA\cite{antol2015vqa}              &ResNet + ResNeXt                                    & 42.64                              & 46.34                              & 45.82                              & 44.24                            & 43.27                              & 46.93                              & 45.62                              & 44.92                            \\
STVQA\cite{jang2017tgif}             &ResNet + ResNeXt                                    & 44.76                              & 49.26                              & 55.86                              & 47.94                            & 45.51                              & 47.57                              & 54.59                              & 47.64                            \\
CoMem\cite{gao2018motion}            &ResNet + ResNeXt                                     & 45.22                              & 49.07                              & 55.34                              & 48.04                            & 45.85                              & 50.02                              & 54.38                              & 48.54                            \\
HME\cite{fan2019heterogeneous}               &ResNet + ResNeXt                                    & 46.18                              & 48.20                              & 58.30                              & 48.72                            & 46.76                              & 48.89                              & 57.37                              & 49.16                            \\
HCRN\cite{le2020hierarchical}             &ResNet + ResNeXt                                     & 45.91                              & 49.26                              & 53.67                              & 48.20                            & 47.07                              & 49.27                              & 54.02                              & 48.89                            \\
HGA\cite{jiang2020reasoning}              &ResNet + ResNeXt                                     & 46.26                              & 50.74                              & 59.33                              & 49.74                            & 48.13                              & 49.08                              & 57.79                              & 50.01                            \\
IGV\cite{li2022invariant}              &ResNet + ResNeXt                                     & -                                  & -                                  & -                                  & -                                & 48.56                              & 51.67                              & 59.64                              & 51.34                            \\
HQGA\cite{xiao2022video1}             &ResNet + ResNeXt + FasterRCNN                                     & 48.48                              & 51.24                              & 61.65                              & 51.42                            & 49.04                              & \textbf{52.28}                     & 59.43                              & 51.75                            \\
ATP\cite{buch2022revisiting}              &CLIP                                     & 53.1                               & 50.2                               & \textbf{66.8}                      & 54.3                             & -                                  & -                                  & -                                  & -                                \\
VGT\cite{xiao2022video}             &ResNet + ResNeXt + FasterRCNN                                      & 52.28                              & \underline{55.09}                              & \underline{64.09}                              & \underline{55.02}                            & 51.62                              & 51.94                              & \textbf{63.65}                     & 53.68                            \\
EIGV\cite{li2022equivariant}             &ResNet + ResNeXt                                     & -                                  & -                                  & -                                  & -                                   & -                                  & -                                  & -                                  & \underline{53.7}                             \\ \midrule
\textbf{VCSR-ResNet*}      &ResNet + ResNeXt    & 50.17                              & 50.74                              & 57.92                              & 51.56                            & 49.62                              & 50.28                              & 61.00                              & 51.69                            \\
\textbf{VCSR-ResNet}     &ResNet + ResNeXt                              & 50.9                               & 51.3                               & 58.36                              & 52.22                            & 49.98                             & \underline{51.98}                              & 61.78                              & 52.53                           \\
\textbf{VCSR-CLIP*}        &CLIP          & \underline{53.13}                              & 53.23                              & 62.55                              & 54.62                            & \underline{52.00}                              & 50.88                              & 60.64                              & 53.07                            \\
\textbf{VCSR-CLIP}        &CLIP                                & \textbf{54.12}                     & \textbf{55.33}                     & 63.06                              & \textbf{55.92}                   & \textbf{53.00}                     & 51.52                              & \underline{62.28}                              & \textbf{54.06}                   \\ \bottomrule
\end{tabular}
}
\setlength{\abovecaptionskip}{0.5mm}
\caption{Comparison with state-of-the-art methods on NExT-QA dataset. The \textbf{best} and \underline{second-best} results are highlighted. The ``\textbf{VCSR-ResNet*}'' and ``\textbf{VCSR-CLIP*} denote the VCSR models that do not incorporate QGR and CSS modules and are trained without contrastive learning objective $\mathcal{L}_{VC}$ and semantic preserving objective $\mathcal{L}_{SP}$.}
\vspace{-10pt}
\label{table:next}
\end{table*}

\textbf{Segment-frame semantic preserving loss}.
To preserve specific semantics of segment and frame features, we propose a novel segment-frame semantic preserving loss.
This loss is based on the assumption that if a single frame is sufficient to answer a question, then the video segment that contains that frame should also be sufficient for the same question. The above assumption described that the positive video segment should be relatively more important in answering the question. We estimate the relative importance of two types of visual contents with their cosine similarity to the question representation:
\begin{equation}
    I =  [I^i_f, I^i_s] = \textrm{Softmax}([\textrm{sim}(q_g, f^i_p), \textrm{sim}(q_g, s^i_p)])
\end{equation}
where $\textrm{sim}(\cdot)$ refers to the cosine similarity, $[,]$ means concatenation, $I^i_f$ and $I^i_s$ are the relative importance of $i$-th positive frame and segment. We then introduce hinge loss to model the relative importance constraint:
\begin{equation}
    \mathcal{L}_{SP} = \sum^k_{i = 1}\textrm{max}( (I^i_f - I^i_s), 0)
    \label{loss:gc}
\end{equation}

\subsection{Video Question Reasoning}
Given the causal scene sets $C_g = [C_p, C_n]$ (i.e., the intervened scene set $s'$ in eq.\ref{eq: causal explanation1}, by fixing the positive scene set, we implement $do(P = p)$), we leverage contrastive learning to model the reasoning about causal interventions based on scene separating. As shown in Figure.\ref{fig:qa reasoner}, we derive answer representations by feeding the multi-modal transformer (MMT) reasoner with positive, negative, and global causal scenes:
\begin{equation}
    a_p = MMT(ME(C_p), ME(q_l))
\end{equation}
\begin{equation}
    a_n = MMT(ME(C_n), ME(q_l))
\end{equation}
\begin{equation}
    a_g = MMT(ME(C_g), ME(q_l))
\end{equation}
where $ME$ is modality embedding module ,$a_p$ and $a_n$ are answer contrastive counterparts, and $a_g$ acts as the contrastive anchor.

\textbf{Visual contrastive loss}.
To estimate the causal effect of the scene-separating intervention, we introduce InfoNCE loss to construct a contrastive objective as follows:

\begin{equation}
    \mathcal{L}_{VC} = -\log\frac{e^{a^T_p\cdot a_g}}{e^{a^T_p \cdot a_g} + \sum^{\mathcal{N}}_{i = 1} e^{a^T_p \cdot a^i_g}}
    \label{loss:vc}
\end{equation}
where $\mathcal{N}$ is the number of negative answers, those answers are obtained by feeding the QA reasoner with different sampling subsets of negative scenes.

\subsection{Answer Prediction}
For multi-choice QA settings, the local question representation $q_l$ is derived by feeding the concatenation of questions and answer candidates to the question encoder. And the answer prediction is given by the positive part of answer representations:
\begin{equation}
    \widetilde{a} = \arg\max(F(a_p))
\end{equation}
where $F$ is a set of linear projections that $F = \{f_a\}^\mathcal{|A|}_{a=1}$, $\mathcal{A}$ is the set of answer candidates, $f_a \in \mathbb{R}^{d \times 1}$ denotes the final linear head for each question candidates.

As for the open-ended QA setting, the formulation of the final answer prediction is:
\begin{equation}
    \widetilde{a} = \arg\max(f_o(a_p))
\end{equation}
in which $f_o \in \mathbb{R}^{d \times \mathcal{|A|}}$ is a fully-connected layer, and $\mathcal{|A|}$ denotes the length of answer dictionary.

\textbf{Question answering loss}. The question-answering loss is the cross entropy loss between the predicted answer $\widetilde{a}$ and the ground truth answer $a_{gt}$:
\begin{equation}
    \mathcal{L}_{QA} = \textrm{CrossEntropy}(\widetilde{a}, a_{gt})
    \label{loss:qa}
\end{equation}

\subsection{Training objective}
Our total training objective comprises three components: question-answering loss (See eq.\ref{loss:qa}), visual contrastive loss (See eq.\ref{loss:vc}), and segment-frame semantic preserving loss (See eq.\ref{loss:gc}), the overall objective is achieved by aggregating the above three objectives:
\begin{equation}
\vspace{-5pt}
    \mathcal{L} = \mathcal{L}_{QA} + \alpha \mathcal{L}_{VC} + \beta \mathcal{L}_{SP}
\end{equation}
where $\alpha$ and $\beta$ are hyper-parameters that control the contribution of sub-objectives.


\begin{table*}
\resizebox{\textwidth}{!}{
\begin{tabular}{lccccccccc}
\hline
\multirow{2}{*}{\textbf{Methods}}                      & \multirow{2}{*}{\textbf{$Acc_E$}} & \multirow{2}{*}{\textbf{$Acc_D$}} & \multicolumn{3}{c}{\textbf{$Acc_P$}}                                              & \multicolumn{3}{c}{\textbf{$Acc_C$}}                                              & \multirow{2}{*}{\textbf{$Acc$}}   \\ \cline{4-9}
                                      &                                 &                                 & \textbf{$Q \rightarrow A$} & \textbf{$Q \rightarrow R$} & \textbf{$Q \rightarrow AR$} & \textbf{$Q \rightarrow A$} & \textbf{$Q \rightarrow R$} & \textbf{$Q \rightarrow AR$} &                \\ \hline
EVQA\cite{antol2015vqa}                                  & 60.95                           & 63.73                           & 45.68                    & 46.40                    & 27.19                     & 48.96                    & 51.46                    & 30.19                     & 45.51          \\
CoMem\cite{gao2018motion}                                 & 62.79                           & 64.08                           & 51.00                    & 50.36                    & 31.41                     & 51.61                    & 53.10                    & 32.55                     & 47.71          \\
HME\cite{fan2019heterogeneous}                                   & 61.45                           & 63.36                           & 50.29                    & 47.56                    & 28.92                     & 50.38                    & 51.65                    & 30.93                     & 46.16          \\
HCRN\cite{le2020hierarchical}                                  & 61.61                           & 65.35                           & 51.74                    & 51.26                    & 32.57                     & 51.57                    & 53.44                    & 32.66                     & 48.05          \\
HGA\cite{jiang2020reasoning}                                   & 63.51                           & 65.67                           & 49.36                    & 50.62                    & 32.22                     & 52.44                    & \underline{55.85}                    & \underline{34.28}                     & 48.92          \\
B2A\cite{park2021bridge}                                   & 62.92                           & \textbf{66.21}                  & 48.96                    & 50.22                    & 31.15                     & \underline{53.27}           & \textbf{56.27}           & \textbf{35.16}            & 49.11          \\ \hline
\textbf{VCSR-CLIP*} & \underline{64.91}                           & 65.00                           & \underline{57.69}                    & \underline{54.74}                    & \underline{36.74}                     & 52.26                    & 53.14                    & 32.27                     & \underline{49.73}          \\
\textbf{VCSR-CLIP}                        & \textbf{65.41}\textbf{(+0.5)}                  & \underline{65.98}\textbf{(+0.98)}                           & \textbf{60.88}\textbf{(+3.19)}           & \textbf{58.54}\textbf{(+3.8)}           & \textbf{41.24}\textbf{(+4.5)}            & \textbf{53.38}\textbf{(+1.12)}                    & 54.37\textbf{(+1.23)}                    & 34.06\textbf{(+1.79)}                     & \textbf{51.67}\textbf{(+1.94)} \\ \hline
\end{tabular}
}
\setlength{\abovecaptionskip}{0.5mm}
\caption{Comparison with state-of-the-art methods on Causal-VidQA dataset. ($E$: explanatory, $D$: descriptive, $P$: prediction, $C$: counterfactual, $Q$: question, $A$: answer, $R$: reason)}
\vspace{-10pt}
\label{table:causal}
\end{table*}

\vspace*{-5pt}
\section{Experiments}
\subsection{Datasets}
We evaluate VCSR on three VideoQA benchmarks that evaluate the model's reasoning capacity from different aspects including temporality, causality, and commonsense: \textbf{NExT-QA}\cite{xiao2021next}, \textbf{Causal-VidQA}\cite{li2022representation} and \textbf{MSRVTT-QA}\cite{xu2017video}.

\textbf{NExT-QA} highlights the causal and temporal relations among objects in videos. It is a manually annotated multi-choice QA dataset targeting the explanation of video contents, especially causal and temporal reasoning. It contains 5,440 videos and 47,692 QA pairs, each QA pair comprises one question and five candidate answers.

\textbf{Causal-VidQA} emphasizes both evidence reasoning and commonsense reasoning in real-world actions. It is a multi-choice QA benchmark containing 107,600 QA pairs and 26,900 video clips. Questions in Causal-VidQA dataset are categorized into four question types: description, explanatory, prediction, and counterfactual. For prediction and counterfactual questions, Causal-VidQA proposed three types of reasoning tasks: question to answer ($Q\rightarrow A$), question to reason ($Q\rightarrow R$), and question to answer and reason ($Q \rightarrow AR$).

\textbf{MSRVTT-QA} focuses on the
visual scene-sensing ability by asking the descriptive questions. It is an open-ended QA benchmark containing 10,000 trimmed video clips and 243,680 QA pairs, with challenges including description and recognition capabilities.
\begin{table}
\begin{tabular}{@{}lcccc@{}}
\toprule
\textbf{Methods}   & \multicolumn{1}{l}{\textbf{What}} & \multicolumn{1}{l}{\textbf{Who}} & \multicolumn{1}{l}{\textbf{How}} & \multicolumn{1}{l}{\textbf{Total}} \\ \midrule
QueST\cite{jiang2020divide}     & 27.9                              & 45.6                             & 83.0                             & 34.6                               \\
HGA\cite{jiang2020reasoning}       & 29.2                              & 45.7                             & 83.5                             & 35.5                               \\
DualVGR\cite{wang2021dualvgr}   & 29.4                              & 45.5                             & 79.7                             & 35.5                               \\
HCRN\cite{le2020hierarchical}      & -                                 & -                                & -                                & 35.6                               \\
QESAL\cite{liu2021question}     & 30.7                              & 46.0                             & 82.4                             & 36.7                               \\
B2A\cite{park2021bridge}    & -                                 & -                                & -                                & 36.9                               \\
ClipBert\cite{lei2021less}  & -                                 & -                                & -                                & 37.4                               \\
ASTG\cite{jin2021adaptive}      & 31.1                              & 48.5                             & 83.1                             & 37.6                               \\
IGV\cite{li2022invariant}       & -                                 & -                                & -                                & 38.3                               \\
HQGA\cite{xiao2022video1}      & -                                 & -                                & -                                & 38.6\\
 \midrule
\textbf{VCSR-CLIP} & \textbf{31.9}                             & \textbf{51.0}                            & \textbf{85.0}                            & \textbf{38.9}                              \\ \bottomrule
\end{tabular}
\caption{Comparison with SOTAs on MSRVTT.}
\vspace{-25pt}
\label{table:msrvtt}
\end{table}

\subsection{Implementation details}
For each video, we uniformly sample 64 frames following \cite{le2020hierarchical}, and extract the features using a pre-trained CLIP (ViT-L/14) encoder. For the questions, we obtain word embeddings using a pre-trained BERT model. For Causal-VidQA dataset, we follow \cite{li2022representation} to add the BERT representation with  Faster-RCNN\cite{ren2015faster} extracted instance representation for a fair comparison.  The model hidden dimension $d$ is set to 512, the segment length $m$ is set to 6, and the positive segment number $k$ is set to 4 for each dataset. The number of MHSA layers $L$ in QGR is set to 2, and the MMT in the video question reasoner is implemented by a 3-layer transformer. The number of heads of all multi-head attention modules is set to 8. The training process is optimized by the AdamW\cite{loshchilov2017decoupled} optimizer with the learning rate $lr=1e-5$, $\beta_1 = 0.9$, $\beta_2 = 0.99$, and weight decay of 0. The hyper-parameter $\alpha$ is set to 0.0125 and $\beta$ is set to 0.04. The training progress is carried out for 50 epochs, and the learning rate is halved if the validation accuracy does not improve after 5 epochs.

\subsection{Comparision with SOTA Methods}
Table \ref{table:next} presents a comparison of our VCSR methods with state-of-the-art (SOTA) methods on the NExT-QA dataset. The results demonstrate that our VCSR achieves superior performance on both the validation set and test set. Notably, our VCSR excels in \emph{Causal} question splits, with an accuracy improvement of 1.02\% and 1.38\% in the validation set and test set, respectively, indicating a stronger causal relational reasoning ability. Additionally, our VCSR achieves competitive performance for \emph{Temporal} questions. This validates that our VCSR can effectively discover temporally sensitive visual scenes in videos. For \emph{Descriptive} questions, our VCSR achieves lower performance than previous methods ATP and VGT. This is because VGT adopts object detection pipeline that makes visual scene sensing more fine-grained. And ATP preserves the most representative frame for each video clip at the cost of harming temporal reasoning ability. Although without object detection, our VCSR can outperform these two methods on more challenging problems \emph{Causal} and \emph{Temporal}.  Moreover, we assess the generalization ability of our VCSR on different visual backbones. VCSR-ResNet\cite{he2016deep} replaces the CLIP visual feature with the concatenation of ResNet-101\cite{xie2017aggregated} extracted appearance feature and ResNeXt-101 extracted motion feature.  The results reveal that the introduction of causal scene intervention also enhances the performance of VCSR-ResNet, highlighting the effectiveness of causal scene intervention on different visual backbones.

To further evaluate the evidence reasoning and commonsense reasoning ability of our VCSR in real-world actions, we evaluate the VCSR on the Causal-VidQA dataset, as shown in Table.\ref{table:causal}. Our VCSR achieves a total accuracy of 51.67\%, outperforming the state-of-the-art B2A \cite{park2021bridge} by 2.56\%. Additionally, for predictive and counterfactual tasks, the introduction of causal intervention significantly promotes the performance of VCSR in answering predictive and counterfactual questions, which require better reasoning capability. This highlights the effectiveness of cross-modal causal relational reasoning when addressing these types of questions.

To evaluate the visual scene-sensing ability of our VCSR, we evaluate our VCSR on open-ended descriptive QA dataset MSRVTT. In Table \ref{table:msrvtt}, we compare the performance of VCSR with the state-of-the-art methods on the MSRVTT dataset. The results show that VCSR has good overall performance on the open-ended dataset, particularly for question types ``Who'' and ``How''.

The experimental results in these three large-scale datasets demonstrate  that our VCSR outperforms state-of-the-art methods in terms of comprehensive understanding of visual concepts, temporality, causality, and commonsense within videos. This validates that our VCSR generalizes well across different VideoQA benchmarks.

\begin{figure*}[t]
    \centering
    \includegraphics[width=0.76\textwidth]{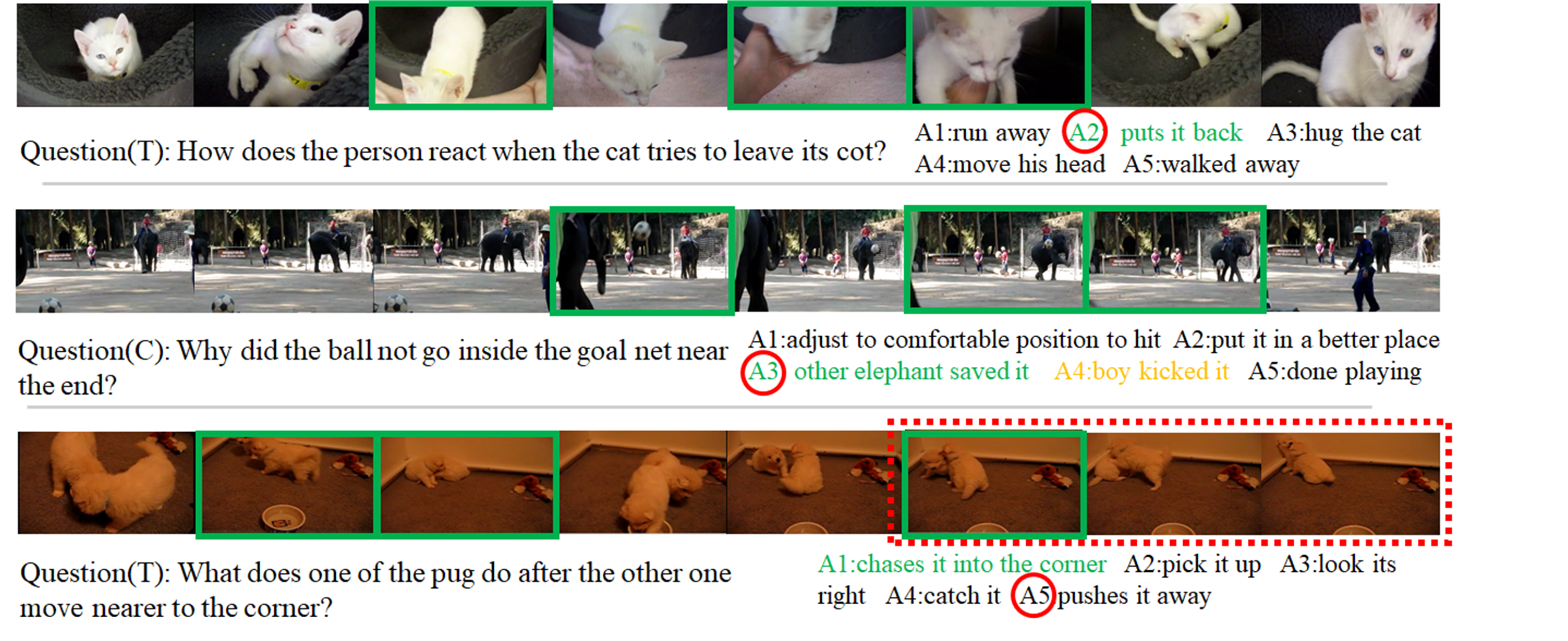}
    \vspace{-15pt}
    \caption{The visualization of causal positive scenes on the NExT-QA dataset. For the first two questions, the positive scenes cover critical video clips and the model predicts the correct answer. However, the model makes wrong answer prediction for the last question, as it cannot fully capture the entire critical scene set. The \textcolor{green}{green} boxes and answers represent the VCSR predicted rationales and answers, respectively, while the \textcolor{red}{red} circles indicate the ground truth answer. In the second example, the \textcolor{orange}{orange} denotes the answer predicted by VCSR*, and in the last example, the \textcolor{red}{red} dashed box shows the human rationale.}
        \vspace{-15pt}
    \label{fig:quality}
\end{figure*}

\vspace{-10pt}
\section{Ablation Studies}
We conduct ablation studies to verify the effectiveness of (1) QGR and CSS module, (2) training objectives $\mathcal{L}_{SP}$ and $\mathcal{L}_{VC}$. All ablation studies are conducted on NExT-QA validation set and MSRVTT-QA dataset, the variants of our VCSR are listed as follows:

\textbf{VCSR-CLIP*}: the VCSR model without QGR and CSS modules and training without contrastive objective $\mathcal{L}_{VC}$ and semantic preserving objective $\mathcal{L}_{SP}$.

\textbf{VCSR-CLIP w/o QGR}: the VCSR model without QGR module, the segment features are obtained by mean-pooling frame features.

\textbf{VCSR-CLIP w/o CSS}: remove the CSS module from VCSR. Without scene separation, the whole scene set is fed to the reasoner. In this setting, the contrastive objective is naturally removed since the lack of counterparts.

\textbf{VCSR-CLIP w/o $\mathcal{L}_{SP}$}: Training without semantic preserving loss  $\mathcal{L}_{SP}$.

\textbf{VCSR-CLIP w/o $\mathcal{L}_{VC}$}: Remove the contrastive objective $\mathcal{L}_{VC}$ from the total objective $L$, the answer prediction is predicted based on the positive answer embedding.

\begin{table}
\resizebox{0.48\textwidth}{!}{
\begin{tabular}{lccccc}
\hline
\multirow{2}{*}{\textbf{Methods}}         & \multicolumn{4}{c}{\textbf{NExT-QA Val}}                                                                                & \multirow{2}{*}{\textbf{MSRVTT-QA}} \\ \cline{2-5}
                                          & \multicolumn{1}{l}{Causal} & \multicolumn{1}{l}{Temporal} & \multicolumn{1}{l}{Descriptive} & \multicolumn{1}{l}{Total} &                                     \\ \hline
\textbf{VCSR-CLIP*}                       & 53.13                      & 53.23                        & 62.55                           & 54.62                     & 38.5                                \\ \hline
\textbf{VCSR-CLIP w/o QGR}                 &52.78      &56.08      &60.49                           & 55.04                     & 38.7                                \\
\textbf{VCSR-CLIP w/o CSS}                & 52.78                      & 54.40                        & 63.35                           & 55.06                     & 38.7                                \\ \hline
\textbf{VCSR-CLIP w/o $\mathcal{L}_{SP}$} & 53.43                      & 54.34                        & 63.19                           & 55.24                     & 38.5                                \\
\textbf{VCSR-CLIP w/o $\mathcal{L}_{VC}$} &53.36      &54.28      &63.06                           & 55.16                     & 38.8                                \\ \hline
\textbf{VCSR-CLIP}                        & 54.12                      & 55.33                        & 63.06                           & 55.92                     & 38.9                                \\ \hline
\end{tabular}
}
\caption{Ablation study on modules and objectives.}
\label{abl:modules}
\vspace{-25pt}
\end{table}

Table \ref{abl:modules} presents the ablation results, indicating that all of the modules and objectives contribute to improving the total performance on both datasets. Specifically, on the validation set of NExT-QA, we observed that removing all modules and objectives would negatively affect the performance of the \emph{Causal} split. Removing QGR, on the other hand, resulted in a decline in the performance of \emph{Causal} and \emph{Descriptive} splits but a boost in the performance of \emph{Temporal} split. This is because the QGR module refining the question-related frames by weighing down other frames in a segment and leading to the partial loss of temporal information.

Moreover, we notice that removing CSS modules, $\mathcal{L}_{SP}$ or $\mathcal{L}_{VC}$, has little effect on the performance of the Descriptive split. For the Descriptive question type, the VCSR-CLIP w/o CSS, w/o $\mathcal{L}_{SP}$, and w/o $\mathcal{L}_{VC}$ are better than the full model VCSR-CLIP in NEXT-QA val. This is because the NEXT-QA dataset is explicitly designed to promote temporal and causal understanding. However, it is important to note that for descriptive question types that emphasize denoised frame-level representation, spatial scene understanding, specific fine-grained spatial information, such as background or salient objects, may be overlooked when focusing on temporal causal scene discovery and semantics preservation. Nonetheless, our proposed CSS, SP, and VC modules significantly contribute to the VCSR model, particularly for causal and temporal question types. Importantly, our VCSR model demonstrates promising performance across all question types, as indicated in the "Total" column. Furthermore, in the MSRVTT-QA dataset, which emphasizes the visual scene-sensing ability through descriptive questions. This confirms the significance of our proposed modules in addressing descriptive questions in relevant datasets.

\section{Qualitative Results}
To verify the ability of the VCSR in discovering visual causal scenes and visual-linguistic causal reasoning, we analyze correct and incorrect visualizations on the NExT-QA dataset. The results are presented in Figure \ref{fig:quality}. When answering the first two questions, the positive scene given by CSS could evidently explain the reason for choosing the correct answer (i.e., scenes of person putting the cat back to the cot and elephant saving the ball). This validates that the VCSR can reliably focus on the dominant visual scenes when making decisions. For the second question, we compare the answer predicted by VCSR and VCSR* and find that the VCSR* without causal intervention is affected by a spurious correlation between visual content ``boy'' and ``ball'', leading to the wrong answer of ``boy kicked it".  In our VCSR, we reduce
such spurious correlation by adopting causal intervention, resulting in better dominant visual evidence and question intention. Moreover, we observe that when answering the last question, the CSS does not capture the entire causal scene set and thus predicts the wrong answer. This is probably caused by the similarity of the visual semantics of the pug's actions, which could be addressed with better visual backbones.

\section{Conclusion}
In this paper, we propose a cross-modal causal relational reasoning framework named VCSR for VideoQA, to explicitly discover the visual causal scenes through causal front-door interventions. From the perspective of causality, we model the causal effect between video-question pairs and the answer based on the structural causal model (SCM). To obtain representative segment features for front-door intervention, we introduce the Question-Guided Refiner (QGR) module. To identify visual causal and non-causal scenes, we propose the Causal Scene Separator (CSS) module. Extensive experiments on three benchmarks demonstrate the superiority of VCSR over the state-of-the-art methods. We believe our work could inspire more causal analysis research in vision-language tasks.
\begin{acks}
This work is supported by the National Key R\&D Program
of China under Grant 2021ZD0111601, in part by the National Natural Science Foundation of China under Grants 62002395 and 61976250, in part by
the Guangdong Basic and Applied Basic Research Foundation under Grants
2023A1515011530, 2021A1515012311, and 2020B15\\15020048, and in part
by the Guangzhou Science and Technology Planning Project under Grant 2023A04J2030.
\end{acks}



\bibliographystyle{ACM-Reference-Format}
\balance
\bibliography{acmart}

\end{document}